\documentclass[letterpaper, 12pt]{ewshm}

\usepackage[dvips,final]{graphicx}

\usepackage[table]{xcolor}
\usepackage{makecell}
\usepackage{enumerate}
\usepackage{color}
\usepackage{subfig}
\usepackage{caption}
\usepackage{threeparttable}
\usepackage{amsfonts}
\usepackage{amsmath}
\usepackage{amssymb}
\usepackage{times}
\usepackage{cite}
\usepackage{hhline}
\usepackage{compactbib}
\usepackage{graphicx}        
\usepackage{amsmath,amsfonts,amssymb}
\usepackage{geometry}
\usepackage[labelsep=period]{caption}

\definecolor{halfgreen}{RGB}{0,128,0}
\definecolor{ahsred}{RGB}{192,0,0}

\renewcommand{\thefigure}{\arabic{figure}}
\renewcommand{\thetable}{\Roman{table}}

\newcommand{\beq}{\begin{equation}}
\newcommand{\eeq}{\end{equation}}
\newcommand{\bgqar}{\begin{eqnarray}}
\newcommand{\enqar}{\end{eqnarray}}
\newcommand{\bgqarn}{\begin{eqnarray*}}
\newcommand{\enqarn}{\end{eqnarray*}}
\newcommand{\bgary}{\begin{array}}
\newcommand{\enary}{\end{array}}

\long\def\symbolfootnote[#1]#2{\begingroup%
\def\thefootnote{\fnsymbol{footnote}}\footnote[#1]{#2}\endgroup}

\makeatletter
\renewcommand\@biblabel[1]{#1.}
\makeatother

\begin{document}


\vspace*{4.4cm}

\noindent Title: \textbf{Transformer-Based Indirect Structural Health Monitoring of Rail Infrastructure with Attention-Driven Detection and Localization of Transient Defects}

\vspace{1.6cm}

\noindent
$
\begin{array}{ll}
\text{Authors}: 
& \text{Sizhe Ma}  \\ 
& \text{Katherine A. Flanigan} \\ 
& \text{Mario Berg\'es} \\
& \text{James Brooks} \\

\end{array}
$

\newpage


\vspace*{60mm}

\noindent \uppercase{\textbf{ABSTRACT}} \vspace{12pt} 

Indirect structural health monitoring (iSHM) for broken rail detection using onboard sensors presents a cost-effective paradigm for railway track assessment, yet reliably detecting small, transient anomalies (2-10 cm) remains a significant challenge due to complex vehicle dynamics, signal noise, and the scarcity of labeled data limiting supervised approaches. This study addresses these issues through unsupervised deep learning. We introduce an incremental synthetic data benchmark designed to systematically evaluate model robustness against progressively complex challenges like speed variations, multi-channel inputs, and realistic noise patterns encountered in iSHM. Using this benchmark, we evaluate several established unsupervised models alongside our proposed Attention-Focused Transformer. Our model employs a self-attention mechanism, trained via reconstruction but innovatively deriving anomaly scores primarily from deviations in learned attention weights, aiming for both effectiveness and computational efficiency. Benchmarking results reveal that while transformer-based models generally outperform others, \textit{all} tested models exhibit significant vulnerability to high-frequency localized noise, identifying this as a critical bottleneck for practical deployment. Notably, our proposed model achieves accuracy comparable to the state-of-the-art solution while demonstrating better inference speed. This highlights the crucial need for enhanced noise robustness in future iSHM models and positions our more efficient attention-based approach as a promising foundation for developing practical onboard anomaly detection systems.


\symbolfootnote[0]{\hspace*{-7mm} Sizhe Ma\textsuperscript{1}, Katherine A. Flanigan, Ph.D\textsuperscript{2} (Corresponding author), Mario Berg\'es, PhD\textsuperscript{3} (Mario Berg\'es holds concurrent appointments at Carnegie Mellon University (CMU) and as an Amazon Scholar. This manuscript describes work at CMU and is not associated with Amazon.). Email: \{sizhem\textsuperscript{1}, kflaniga\textsuperscript{2}, mberges\textsuperscript{3}\}@andrew.cmu.edu. Department of Civil and Environmental Engineering, Carnegie Mellon University, Pittsburgh, PA, USA. \\
James Brooks, PhD. Email: brooksjd@gcc.edu. Department of Electrical and Computer Engineering, Grove City College, Grove City, PA, USA.}


\vspace{12pt} 
\noindent \uppercase{\textbf{INTRODUCTION}}  \vspace{12pt} 

Indirect structural health monitoring (iSHM), typically implemented through onboard sensor systems on operational trains, presents a promising paradigm for assessing railway infrastructure health \cite{hoelzl_chapter_2022}. By utilizing sensors such as accelerometers mounted on vehicles, iSHM aims to infer the condition of components like bridges or the track itself without requiring direct instrumentation of the infrastructure \cite{gkoumas_way_2021}. This approach offers potential advantages over traditional direct monitoring methods, including reduced costs associated with sensor installation and maintenance across extensive networks, and the ability to leverage routine train operations for data acquisition. The importance of such monitoring is underscored by its potential to enhance railway safety by detecting precursors to failure and to facilitate more efficient, condition-based maintenance strategies, thereby optimizing resource allocation and improving network availability. However, a significant challenge persists in reliably detecting small-scale (approximately 2-10 cm), transient anomalies directly on the rail track using iSHM \cite{hoelzl_chapter_2022}. These anomalies, which may include short corrugations and transient anomalies at similar lengths (on switches, welds, insulated joints, and squats), can develop rapidly under operational loads and potentially escalate to critical failures. The indirect nature of iSHM measurements, capturing the vehicle's complex dynamic response rather than the defect itself, combined with the subtle signature of these small, transient events, makes their detection particularly difficult amidst background noise and operational variability, even in lab settings \cite{montero_anomaly_2023,yin_2023}. 

Machine learning techniques offer powerful capabilities for analyzing the large volumes of time-series data generated by onboard monitoring \cite{shim_anomaly_2022}. However, the application of conventional supervised machine learning is significantly hampered by the difficulty in obtaining comprehensive, accurately labeled ground truth data for transient railway anomalies. Creating datasets that exhaustively identify small-scale defects or certify long track segments as completely defect-free across extensive networks is considered prohibitively expensive and practically infeasible. The scarcity of reliable labels limits the effectiveness of supervised approaches, naturally leading to increased interest in unsupervised methods. Given that onboard sensor data is typically high-frequency (thousands of Hz), multi-channel, and exhibits complex, nonlinear temporal patterns reflecting vehicle-track dynamics, unsupervised deep learning is particularly well-suited \cite{hoelzl_chapter_2022}. Architectures like autoencoders and recurrent neural network can automatically learn representative features and temporal dependencies from large amounts of unlabeled, high-dimensional sequential data without requiring manual feature engineering. 

Despite this theoretical suitability, unsupervised deep learning solutions capable of achieving the accuracy and consistency demanded for industrial deployment in detecting transient track anomalies are still lacking. A primary challenge remains the difficulty in rigorously validating model performance and reliability due to the absence of comprehensive ground truth labels. Furthermore, the inherent complexity and noise within onboard sensor signals---influenced by high-frequency vibrations, speed variations, vehicle dynamics, and environmental factors---make it difficult for models to reliably distinguish subtle anomalies from normal operational variability, potentially leading to high false alarm rates \cite{li_overview_2017}. Finally, the inherent ``black-box'' characteristic of deep learning models makes it challenging to understand their internal decision-making processes, particularly \textit{why} they misinterpret complex signal features or noise as anomalies, thus impeding the diagnosis and mitigation of high false alarm rates in safety-critical applications. 

To systematically investigate the limitations of current unsupervised deep learning approaches and pinpoint the specific factors hindering their reliable application in railway iSHM, this study introduces an incremental synthetic data benchmark. This methodology allows for controlled testing where signal complexity is gradually increased, simulating challenges such as multi-channel inputs, speed variations, and high-frequency noise commonly encountered in real-world accelerometer data. By evaluating the performance of several state-of-the-art unsupervised deep learning algorithms against this benchmark, we can identify the specific conditions under which these models falter, thereby revealing critical bottlenecks, such as sensitivity to noise, that impede robust transient anomaly detection. Alongside this benchmarking effort, we propose and evaluate an advanced unsupervised deep learning model incorporating attention mechanisms. While our proposed model does not outperform all benchmarked models in raw accuracy across every scenario (although it remains comparable), it achieves a compelling balance between detection performance and computational efficiency---an essential requirement for indirect broken rail detection, where anomalies must be identified before the next train passes. This favorable trade-off positions it as a strong candidate for practical onboard deployment and provides a solid foundation for future research focused on enhancing robustness against the specific challenges, particularly high-frequency noise, identified through our systematic evaluation.


\vspace{20pt}

\noindent \uppercase{\textbf{Incremental synthetic data benchmark}} \vspace{12pt}

\begin{figure}[!t] 
  \centering{\includegraphics[width=1\columnwidth]{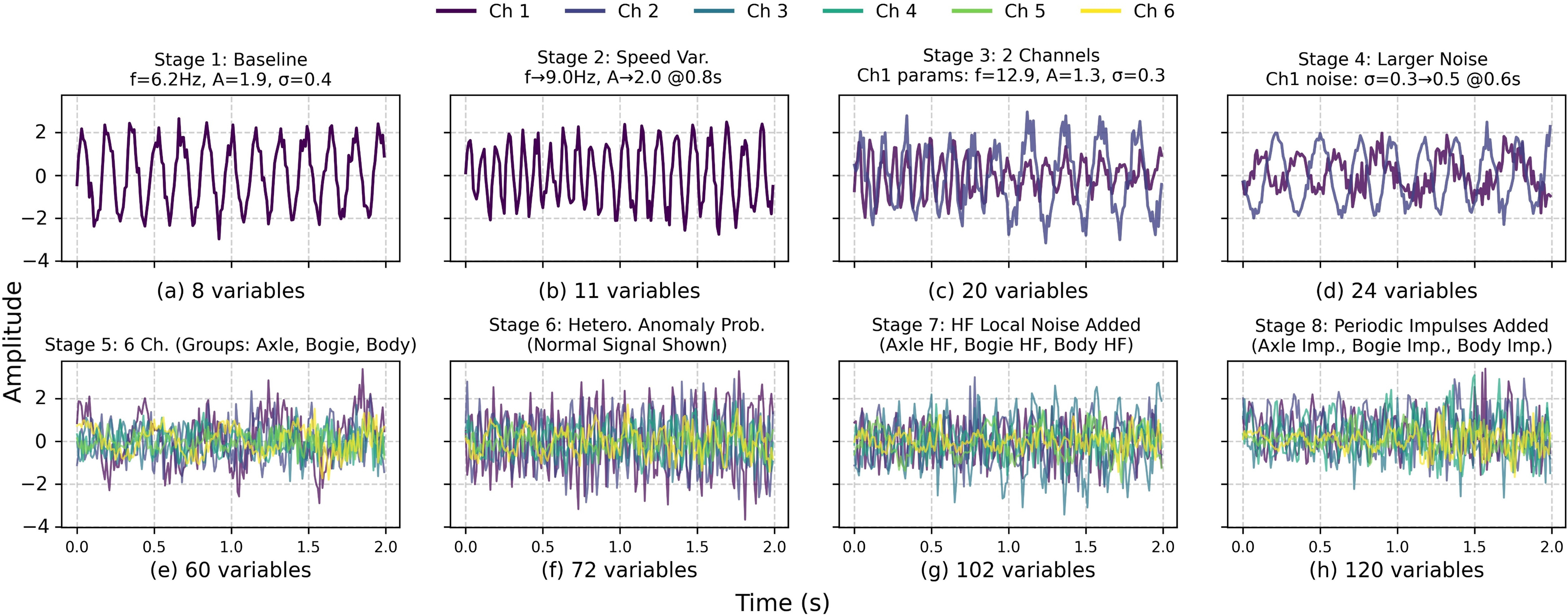}}
    \vspace{-0.7cm}
    \caption{\small Time-series signals for eight benchmarking stages with increasing complexity: speed variation, multi-channel inputs, varying noise levels, localized high-frequency noise, and periodic impulses. Each plot shows the \# of underlying random variables.}
    \label{fig:benchmark_stages} \vspace{-0.3cm}
\end{figure}

Our motivation for developing an incremental synthetic data benchmark stems from challenges encountered when applying state-of-the-art unsupervised anomaly detection models to real-world railway iSHM data. We have collected extensive datasets using six accelerometers (two each on axle boxes, bogie frames, and the car body) sampled at 2000 Hz, yet achieving consistent and accurate detection of small, transient rail anomalies proved difficult. To systematically investigate the factors limiting model performance under increasing signal complexity, this benchmark was created, as shown in Figure~\ref{fig:benchmark_stages}. It progressively introduces complexities analogous to those observed in real onboard sensor data, enabling controlled assessment of where different models falter. While leveraging parameter insights from literature \cite{hoelzl_chapter_2022, montero_anomaly_2023}, the benchmark's design aims to retain maximal physical relevance to our acquisition setup. Specifically, each synthetic signal instance spans 2 seconds, mirroring the window length chosen for our real-world data analysis, which reflects the theoretical maximum time required for our instrumented vehicle to fully traverse target transient anomalies. We generate these synthetic signals at 100 Hz; this deliberate choice establishes a $\times$20 scaling factor for time and frequency-related parameters compared to our 2000 Hz field data, allowing exploration of fundamental dynamics at a reduced computational scale. Amplitude parameters are chosen within plausible ranges intended to pose non-trivial detection challenges. For each generated instance, signal parameters are independently sampled from specified uniform distributions $U[a,b]$ or Gaussian distributions $\mathcal{N}(\mu, \sigma^2)$.

\vspace{12pt}
\noindent\textbf{Stage 1: Baseline Single-Channel Signal.}
Establishes a baseline single-channel signal over $t \in [0, 2]$s:
\begin{equation*}
    s(t) = A \sin(2\pi f t) + n(t)
\end{equation*}
Parameters are frequency $f\in U[1, 15]$Hz, amplitude $A \in U[0.5, 2.0]$, and additive white noise $n(t) \sim \mathcal{N}(0, \sigma^2)$ with standard deviation $\sigma \in U[0.1, 0.5]$. Anomalies occur in 10\% of instances (one per instance): 5\% are instantaneous spikes (at random $t_{spike} \in U[0,2]$s, offset $\delta_{spike} \in \pm U[2, 4]$), 5\% are local constant deviations (duration $\Delta t_{local} \in U[0.05, 0.2]$s, start $t_{local,start} \in U[0, 2 - \Delta t_{local}]$s, offset $\delta_{local} \in \pm U[1, 2]$).

\vspace{12pt}
\noindent\textbf{Stage 2: Speed Variation via Frequency Change.}
Introduces non-stationarity via a frequency and amplitude shift at $t_{change}$, maintaining phase continuity:
\begin{equation*}
    s(t) = 
    \begin{cases} 
      A \sin(2\pi f t) + n(t), & t < t_{change} \\
      k_A A \sin[2\pi (k_f f) (t \!-\! t_{change}) + 2\pi f t_{change}] + n(t), & t \ge t_{change}
    \end{cases}
\end{equation*}
Speed variation parameters are change time $t_{change} \in U[0.4, 1.6]$s, frequency factor $k_f \in U[0.5, 1.5]$, and amplitude factor $k_A \in U[0.5, 1.5]$. Anomaly injection follows Stage 1.

\vspace{12pt}
\noindent\textbf{Stage 3: Coupled Multi-Sensor Signals.}
Expands to two channels ($i=1, 2$) with independent parameters. Each channel $s_i(t)$ follows the piecewise model of Stage 2, incorporating a channel-specific phase offset $\phi_i$:
\begin{equation*}
    s_i(t) = \!
    \begin{cases} 
      A_i \sin(2\pi f_i t \!+\! \phi_i) \!+\! n_i(t), & \!\! t < t_{change,i} \\
      k_{A,i} A_i \sin(2\pi (k_{f,i} f_i) (t \!-\! t_{change,i}) \!+\! 2\pi f_i t_{change,i} \!+\! \phi_i) \!+\! n_i(t), & \!\! t \ge t_{change,i}
    \end{cases}
\end{equation*}
Each channel has its own parameters $\{A_i, f_i, \sigma_i, t_{change,i}, k_{f,i}, k_{A,i}\}$ sampled independently using ranges from Stages 1 and 2. The phase offset $\phi_i \in U[0, 2\pi]$ is also sampled independently per channel. Noise $n_i(t)$ is independent per channel. Anomalies (10\% total rate) affect only one channel per instance, chosen uniformly randomly.

\vspace{12pt}
\noindent\textbf{Stage 4: Increased Noise Variability.}
Introduces piecewise noise levels per channel. Let $S_{3,i}(t)$ be the signal for channel $i$ from Stage 3. The signal $s_i(t)$ is:
\begin{equation*}
    s_i(t) = S_{3,i}(t) + 
    \begin{cases} 
      n_i^{(1)}(t), & t < t_{noisechange,i} \\
      n_i^{(2)}(t), & t \ge t_{noisechange,i}
    \end{cases}
\end{equation*}
The noise standard deviation changes from $\sigma_i$ (sampled as before) to $\sigma_i + \Delta\sigma_i$, with the increase $\Delta\sigma_i \in U[0.1, 0.5]$. The noise terms are $n_i^{(1)}(t) \sim \mathcal{N}(0, \sigma_i^2)$ and $n_i^{(2)}(t) \sim \mathcal{N}(0, (\sigma_i + \Delta\sigma_i)^2)$. Anomaly injection follows Stage 3.

\vspace{12pt}
\noindent\textbf{Stage 5: Six Channels with Different Ranges per Sensor Group.}
Expands to six channels ($i=1...6$) grouped as Axle (channels 1-2), Bogie (channels 3-4), Body (channels 5-6). Each channel follows the Stage 4 model, with all parameters sampled independently per channel. Parameters are sampled from different ranges in different groups. For example, base frequency $f_i$ ranges become group-specific: $U[1, 25]$ Hz (Axle), $U[1, 15]$ Hz (Bogie), $U[1, 5]$ Hz (Body). Anomaly injection follows Stage 3 logic.

\vspace{12pt}
\noindent\textbf{Stage 6: Heterogeneous Anomaly Likelihood Across Channels.}
Same signal generation as Stage 5. Anomaly allocation changes: the affected channel $i$ is chosen non-uniformly based on pre-defined probabilities $p_{spike,i}$ for spikes and $p_{local\_dev,i}$ for local deviations, where $\sum_{i=1}^6 p_{spike,i} = 1$ and $\sum_{i=1}^6 p_{local\_dev,i} = 1$. Anomaly rate is 10\%. 

\vspace{12pt}
\noindent\textbf{Stage 7: High-Frequency Local Noise.}
Adds transient high-frequency bursts to Stage 6 signals. Let $S_{6,i}(t)$ be the signal for channel $i$ from Stage 6. The signal becomes:
\begin{equation*}
    s_i(t) = \!
    \begin{cases} 
      S_{6,i}(t) \!+\! \alpha_{HF,i} \sin(2\pi f_{HF,i} t \!+\! \psi_i), & \!\! t_{HF,start,i} \le t \le t_{HF,end,i} \\
      S_{6,i}(t), & \!\! \text{otherwise}
    \end{cases}
\end{equation*}
Local noise parameters: start time $t_{HF,start,i} \in U[0.4, 1.6]$s, duration $\Delta t_{HF,i} \in U[0.05, 0.2]$s, end time $t_{HF,end,i} = t_{HF,start,i} + \Delta t_{HF,i}$, phase $\psi_i \in U[0, 2\pi]$. HF amplitude $\alpha_{HF,i}$ and frequency $f_{HF,i}$ are group-dependent: Axle ($U[0.5, 2.0]$, $U[25, 50]$Hz), Bogie ($U[0.3, 1.5]$, $U[15, 40]$Hz), Body ($U[0.2, 1.0]$, $U[5, 30]$Hz). Anomaly injection follows Stage 6.

\vspace{12pt}
\noindent\textbf{Stage 8: Periodic Impulses.}
Adds periodic truncated Gaussian impulses $I_i(\tau)$ to Stage 7 signals. Let $S_{7,i}(t)$ be the signal for channel $i$ from Stage 7. The total signal is:
\begin{equation*}
    s_i(t) = S_{7,i}(t) + \sum_{k=0}^{K_i-1} I_i(t - k T_i) 
\end{equation*}
The impulse function $I_i(\tau)$ for channel $i$ is defined as:
\begin{equation*}
    I_i(\tau) = 
    \begin{cases} 
      \beta_i \exp \left[ - \left( \frac{\tau}{\omega_i} \right)^2 \right], & 0 \le \tau \le 3\omega_i \\
      0, & \text{otherwise}
    \end{cases}
\end{equation*}
Periodic impulse parameters per channel $i$: impulse width $\omega_i \in U[0.01, 0.1]$s, $\tau$ is time since impulse start. Number of impulses $K_i = \lfloor 2 / T_i \rfloor$. Impulse period $T_i$ and amplitude $\beta_i$ are group-dependent: Axle ($T_i \in U[0.1, 0.3]$s, $\beta_i \in U[0.5, 2.0]$), Bogie ($T_i \in U[0.15, 0.3]$s, $\beta_i \in U[0.3, 1.5]$), Body ($T_i \in U[0.2, 0.5]$s, $\beta_i \in U[0.2, 1.0]$). Anomaly injection follows Stage 6.


\vspace{20pt}

\noindent \uppercase{\textbf{Proposed Model and Models for Benchmarking}} \vspace{12pt}

\noindent\textbf{Proposed Model: Attention-Focused Transformer.}
Motivated by the need for an efficient yet effective model for iSHM, we propose unsupervised anomaly detection using a transformer-like framework: Input Attention $\rightarrow$ Self-Attention Encoder $\rightarrow$ Multilayer Perceptron (MLP) Decoder. The Input Attention layer weighs temporal saliency. The Self-Attention Encoder (multi-head attention, feed-forward networks) captures complex temporal dependencies and context. An MLP Decoder reconstructs the input.

\begin{table}[t!]
\centering
\caption{\small AUC performance on incremental benchmark stages.} \vspace{-0.2cm}
\label{tab:auc_results} 
\resizebox{\textwidth}{!}{
\begin{tabular}{lcccccc}
\hline
\textbf{Stage} & \textbf{Model} & \textbf{\begin{tabular}{@{}c@{}}LSTM Encoder- \\ Decoder \cite{malhotra_lstm-based_2016}\end{tabular}} & \textbf{\begin{tabular}{@{}c@{}}CNN \\ Autoecoder\end{tabular}} & \textbf{MSCRED \cite{zhang_detecting_2018}} & \textbf{\begin{tabular}{@{}c@{}}Anomaly \\ Transformer \cite{xu_anomaly_2022}\end{tabular}} & \textbf{\begin{tabular}{@{}c@{}}Our Attention- \\ Based Transformer\end{tabular}} \\ \hline
Step 1 & \textit{Baseline} & 0.956 & 0.960 & 0.977 & 0.989 & 0.992 \\ 
Step 2 & \textit{Speed Variation} & 0.942 & 0.921 & 0.969 & 0.989 & 0.988 \\
& (Drop) & (-0.014) & \textcolor{red}{(-0.039)} & (-0.008) & (--) & (-0.004) \\ 
Step 3 & \textit{2 Channels} & 0.911 & 0.905 & 0.960 & 0.988 & 0.989 \\
& (Drop) & \textcolor{red}{(-0.031)} & (-0.016) & (-0.009) & (-0.001) & (--) \\ 
Step 4 & \textit{Larger Noise} & 0.907 & 0.883 & 0.944 & 0.987 & 0.982 \\
& (Drop) & (-0.004) & \textcolor{red}{(-0.022)} & (-0.016) & (-0.001) & (-0.007) \\ 
Step 5 & \textit{6 Channels} & 0.859 & 0.855 & 0.917 & 0.985 & 0.979 \\
& (Drop) & \textcolor{red}{(-0.048)} & (-0.028) & \textcolor{red}{(-0.027)} & (-0.002) & (-0.003) \\ 
Step 6 & \textit{Homogeneous P} & 0.854 & 0.857 & 0.920 & 0.981 & 0.971 \\
& (Drop) & (-0.005) & (--) & (--) & (-0.004) & (-0.008) \\ 
Step 7 & \textit{HF Local Noise} & 0.693 & 0.732 & 0.782 & 0.864 & 0.844 \\
& (Drop) & \textcolor{red}{(-0.161)} & \textcolor{red}{(-0.125)} & \textcolor{red}{(-0.138)} & \textcolor{red}{(-0.117)} & \textcolor{red}{(-0.127)} \\ 
Step 8 & \textit{Periodic Impulse} & 0.667 & 0.684 & 0.773 & 0.851 & 0.815 \\
& (Drop) & \textcolor{red}{(-0.026)} & \textcolor{red}{(-0.048)} & (-0.009) & (-0.013) & \textcolor{red}{(-0.029)} \\ \hline
\end{tabular}%
} 
    \begin{tablenotes}
      \footnotesize
      \item  \textit{Threshold used}: \{0.5\%, 1.0\%, 1.5\%, 2.0\%, 10\%, 20\%, 30\%\}. Red indicates drop $>2\%$.
    \end{tablenotes} \vspace{-0.2cm}
\end{table}

Crucially, training employs an unsupervised reconstruction objective using a dataset containing \textit{both normal operational data and examples of known anomaly types}. This process guides the model, particularly the Self-Attention layer, to learn representations and reconstruction capabilities for the full spectrum of presented data, including anomalies. Anomaly detection during inference, therefore, relies not on identifying unseen patterns, but on differentiating learned representations. Our scoring mechanism primarily analyzes the attention weight distributions generated by the standard Self-Attention layer. The core assumption is that even when the model reconstructs anomalous segments encountered during training, these segments induce statistically different or more pronounced deviations in the learned contextual attention patterns compared to those typically generated by normal data seen in the same training set. Reconstruction error can serve as a supplementary feature. Compared to specialized approaches, which may use different loss structures or attention mechanisms, our method leverages standard, computationally efficient self-attention components, focusing innovation on interpreting the learned attention distributions to distinguish between normal and anomalous patterns previously seen during training, seeking a balance between performance and efficiency for iSHM. In further benchmarking, we evaluate two variants: one detects based on reconstruction error and another based on attention weights distribution. To evaluate performance against the incremental benchmark, we selected several representative unsupervised anomaly detection models from literature, alongside our proposed approach. The selection covers different architectural paradigms applied to time-series data.

\vspace{12pt}
\noindent\textbf{Benchmark Model: LSTM Encoder-Decoder \cite{malhotra_lstm-based_2016}.}
This model employs Long Short-Term Memory (LSTM) networks in an encoder-decoder setup. The encoder compresses the input sequence into a context vector, and the decoder reconstructs it. Trained on normal data, it assumes anomalies will cause poor reconstruction, yielding high error scores. It handles sequential data well but processes sequentially, limiting parallelization.

\vspace{12pt}
\noindent\textbf{Benchmark Model: CNN Autoencoder.}
This model uses Convolutional Neural Networks (CNNs) in an autoencoder structure. Convolutional layers extract local patterns, and transposed convolutions reconstruct the input. It assumes anomalies disrupt learned local spatio-temporal features, leading to high reconstruction error. CNNs are parallelizable but may need specific designs for long-range dependencies.

\vspace{12pt}
\noindent\textbf{Benchmark Model: MSCRED \cite{zhang_detecting_2018}.}
MSCRED uses Convolutional LSTMs to model multivariate time series, generating multi-scale matrices that capture inter-channel correlations and temporal patterns. Convolutional LSTM encoder-decoder reconstructs the matrices, with anomalies indicated by high reconstruction error due to disrupted correlation structures. While effective for multi-channel data, it is computationally intensive.

\vspace{12pt}
\noindent\textbf{Benchmark Model: Anomaly Transformer \cite{xu_anomaly_2022}.}
A state-of-the-art transformer model adapted for anomaly detection uses a transformer encoder backbone but incorporates a specialized anomaly-attention mechanism that explicitly models discrepancies between learned prior associations and instance-specific series associations to calculate anomaly scores. It is powerful but computationally demanding due to the specialized attention.


\begin{table}[t!]
\centering
\caption{\small Inference time (seconds) on testing sets (3000 instances, batch size 32).}  \vspace{-0.2cm}
\label{tab:time_results_revised}
\resizebox{\textwidth}{!}{
\begin{tabular}{lcccccc}
\hline
\textbf{Stage} & \textbf{Model} & \textbf{\begin{tabular}{@{}c@{}}LSTM Encoder- \\ Decoder \cite{malhotra_lstm-based_2016}\end{tabular}} & \textbf{\begin{tabular}{@{}c@{}}CNN \\ Autoecoder\end{tabular}} & \textbf{MSCRED \cite{zhang_detecting_2018}} & \textbf{\begin{tabular}{@{}c@{}}Anomaly \\ Transformer \cite{xu_anomaly_2022}\end{tabular}} & \textbf{\begin{tabular}{@{}c@{}}Our Attention- \\ Based Transformer\end{tabular}} \\ \hline
Step 1 & \textit{1 Channel} & 156.7 & 67.1 & 169.7 & 142.3 & 100.5 \\ \hline
Step 3 & \textit{2 Channels} & 306.8 & 139.7 & 584.8 & 264.2 & 177.2 \\ \hline
Step 5 & \textit{6 Channels} & 941.5 & 403.9 & 2320.3 & 673.5 & 502.5 \\ \hline
\end{tabular}%
}
\end{table}

\vspace{20pt}

\noindent \uppercase{\textbf{Results and Discussion}} \vspace{12pt}

The performance and inference time of all models were evaluated on the incremental benchmark. Results are shown in Table \ref{tab:auc_results} and Table \ref{tab:time_results_revised}. Time results focus on stages with dimensionality changes (1, 2, 6 channels), as this primarily drove computational cost. The benchmark results indicate that Transformer-based models generally achieve higher area under the curve (AUC) scores than LSTM, CNN, and MSCRED approaches as signal complexity increases. While simpler models perform well initially, challenges like speed variation particularly impact the CNN Autoencoder, likely due to its sensitivity to frequency shifts affecting local patterns. Increasing channels significantly degrades LSTM performance, reflecting difficulties in sequential modeling of multi-channel dependencies. Adding more channels further challenges LSTM and MSCRED, while Transformers maintain robustness.

A critical finding emerges at Step 7: \textit{all models} suffer substantial performance drops. This universal difficulty in distinguishing high-frequency noise bursts from true anomalies represents a major bottleneck for applying current unsupervised methods to iSHM, where similar noise characteristics are common. This sensitivity likely hinders reliable detection of small, transient rail defects. The subsequent addition of periodic impulses causes further, smaller degradations. From an efficiency perspective, the CNN Autoencoder is fastest due to convolutional parallelizability, while MSCRED is significantly slower, especially with more channels, owing to its complex signature matrix computations. Notably, our proposed transformer model is considerably faster than the Anomaly Transformer, while achieving comparable, high AUC scores across most stages.


\vspace{20pt}

\noindent \uppercase{\textbf{Summary}} \vspace{12pt}

This study utilized an incremental synthetic benchmark to evaluate unsupervised deep learning models for challenging iSHM-based transient rail anomaly detection. Our findings highlight a critical, universal vulnerability across tested architectures: significant performance degradation when encountering high-frequency localized noise, representing a key barrier to practical deployment. While transformer-based models, including our proposed attention-focused approach, generally excelled at handling other signal complexities, our model specifically demonstrated a compelling trade-off. It achieved near state-of-the-art AUC performance while offering substantially improved computational efficiency compared to the state-of-the-art Anomaly Transformer. This underscores the urgent need for research focused on noise robustness for reliable iSHM and positions our efficient attention-based model as a promising foundation for developing practical, resource-constrained onboard detection systems.

\vspace{24pt}
\noindent \uppercase{\textbf{Acknowledgments}} 

\vspace{12pt}

This work is support by the Federal Railroad Administration and the Pennsylvania Infrastructure Technology Alliance.




\vspace{20pt}

\small 

\bibliographystyle{iwshm}
\bibliography{IWSHM_references}


\end{document}